\documentclass[sigconf]{acmart}
\usepackage{booktabs} 
\usepackage[
  separate-uncertainty = true,
  multi-part-units = repeat
]{siunitx} 
\usepackage{url}  
\usepackage{graphicx}  
\usepackage{xspace}
\usepackage{color}
\usepackage{subfig}
\usepackage{amsthm}
\usepackage{mathtools,amsthm}
\usepackage{multirow}
\usepackage[colorinlistoftodos]{todonotes}
\usepackage{makecell}

\usepackage{diagbox}
\usepackage{float}
\setcopyright{rightsretained}

\renewcommand\footnotetextcopyrightpermission[1]{}

\usepackage{geometry}
\usepackage{enumitem}
\setlist[itemize]{leftmargin=*}
\setlist[enumerate]{leftmargin=*}

\newtheorem{problem}{Problem}

\newcommand{\nop}[1]{}

\newcommand{\methodFont}{\textsl}

\newcommand{\fixedtime}{\methodFont{Fixedtime}\xspace}

\newcommand{\fixedtimeHistory}{\methodFont{Formula}\xspace}

\newcommand{\sotl}{\methodFont{SOTL}\xspace}
\newcommand{\nips}{\methodFont{DRL}\xspace}
\newcommand{\deeplight}{\methodFont{IntelliLight}\xspace}
\newcommand{\deeplightQueue}{\methodFont{LIT}\xspace}
\newcommand{\ours}{\deeplightQueue}
\newcommand{\formula}{\methodFont{Formula}\xspace}
\newcommand{\pg}{\methodFont{PG}\xspace}
\newcommand{\notationFont}{\mathsf}

\newcommand{\queueLength}{\notationFont{L}}
\newcommand{\numOfVehicles}{\notationFont{V}}
\newcommand{\averageWaitingTime}{\notationFont{W}}
\newcommand{\image}{\notationFont{M}}

\newcommand{\delay}{\notationFont{D}}

\newcommand{\expNotation}{\textsl}

\newcommand{\gWEStraight}{\expNotation{WT-ET}\xspace}

\newcommand{\jn}{City A\xspace}
\newcommand{\hz}{City B\xspace}

\acmDOI{10.475/123_4}

\acmISBN{123-4567-24-567/08/06}

\acmConference[]{ACM conference}{}{}
\acmYear{2019}
\copyrightyear{2019}
\acmArticle{4}
\acmPrice{15.00}

\title{Diagnosing Reinforcement Learning for Traffic Signal Control}

\author{Guanjie Zheng$^\dagger$, Xinshi Zang$^\S$, Nan Xu$^\ddagger$, Hua Wei$^\dagger$}
\par
\author{Zhengyao Yu$^\dagger$, Vikash Gayah$^\dagger$, Kai Xu$^\intercal$, Zhenhui Li$^\dagger$}
\affiliation{$^\dagger$Pennsylvania State University, $^\S$ $^\ddagger$Shanghai Jiao Tong Univerisity, $^\intercal$Shanghai Tianrang Intelligent Technology Co., Ltd\\
$^\dagger$\{gjz5038, hzw77, zuy107, vvg104, zul17\}@psu.edu, \\
$^\S$zang-xs@foxmail.com, $^\ddagger$xunannancy@sjtu.edu.cn, $^\intercal$kai.xu@tianrang-inc.com}

\begin{document}

\begin{abstract}
With the increasing availability of traffic data and advance of deep reinforcement learning techniques, there is an emerging trend of employing reinforcement learning (RL) for traffic signal control. A key question for applying RL to traffic signal control is how to define the reward and state. The ultimate objective in traffic signal control is to minimize the travel time, which is difficult to reach directly. Hence, existing studies often define reward as an ad-hoc weighted linear combination of several traffic measures. However, there is no guarantee that the travel time will be optimized with the reward. In addition, recent RL approaches use more complicated state (e.g., image) in order to describe the full traffic situation. However, none of the existing studies has discussed whether such a complex state representation is necessary. This extra complexity may lead to significantly slower learning process but may not necessarily bring significant performance gain. 

In this paper, we propose to re-examine the RL approaches through the lens of classic transportation theory. We ask the following questions: (1) How should we design the reward so that one can guarantee to minimize the travel time? (2) How to design a state representation which is concise yet sufficient to obtain the optimal solution? Our proposed method \ours is theoretically supported by the classic traffic signal control methods in transportation field. \ours has a very simple state and reward design, thus can serve as a building block for future RL approaches to traffic signal control. Extensive experiments on both synthetic and real datasets show that our method significantly outperforms the state-of-the-art traffic signal control methods.
\end{abstract}


\maketitle

\section{Introduction}

Nowadays, the widely-used traffic signal control systems (e.g., SCATS~\cite{scats,scats90} and SCOOT~\cite{hunt1981scoot,hunt1982scoot}) are still based on manually designed traffic signal plans. At the same time, increasing amount of traffic data have been collected from various sources such as GPS-equipped vehicles, navigational systems, and traffic surveillance cameras. How to utilize the rich traffic data to better optimize our traffic signal control system has drawn increasing attention~\cite{wang2018enhancing}.

The ultimate objective in most urban transportation systems is to  minimize travel time for all the vehicles. With respect to traffic signal control, traditional transportation research typically formulates signal timing as an optimization problem. A key weakness of this approach is that unrealistic assumptions (e.g., uniform arrival rate of traffic or unlimited lane capacity) need to be made in order to make the optimization problem tractable~\cite{Roess2011t}.

On the other hand, with the recent advance of reinforcement learning (RL) techniques, researchers have started to tackle traffic signal control problem through trial-and-error search~\cite{VaOl16, wei2018intellilight}. Compared with traditional transportation methods, RL technique avoids making strong assumption about the traffic and learns directly from the feedback by trying different strategies. 

\begin{figure}
\begin{tabular}{cc}
\hspace{-3mm}\includegraphics[width=0.24\textwidth]{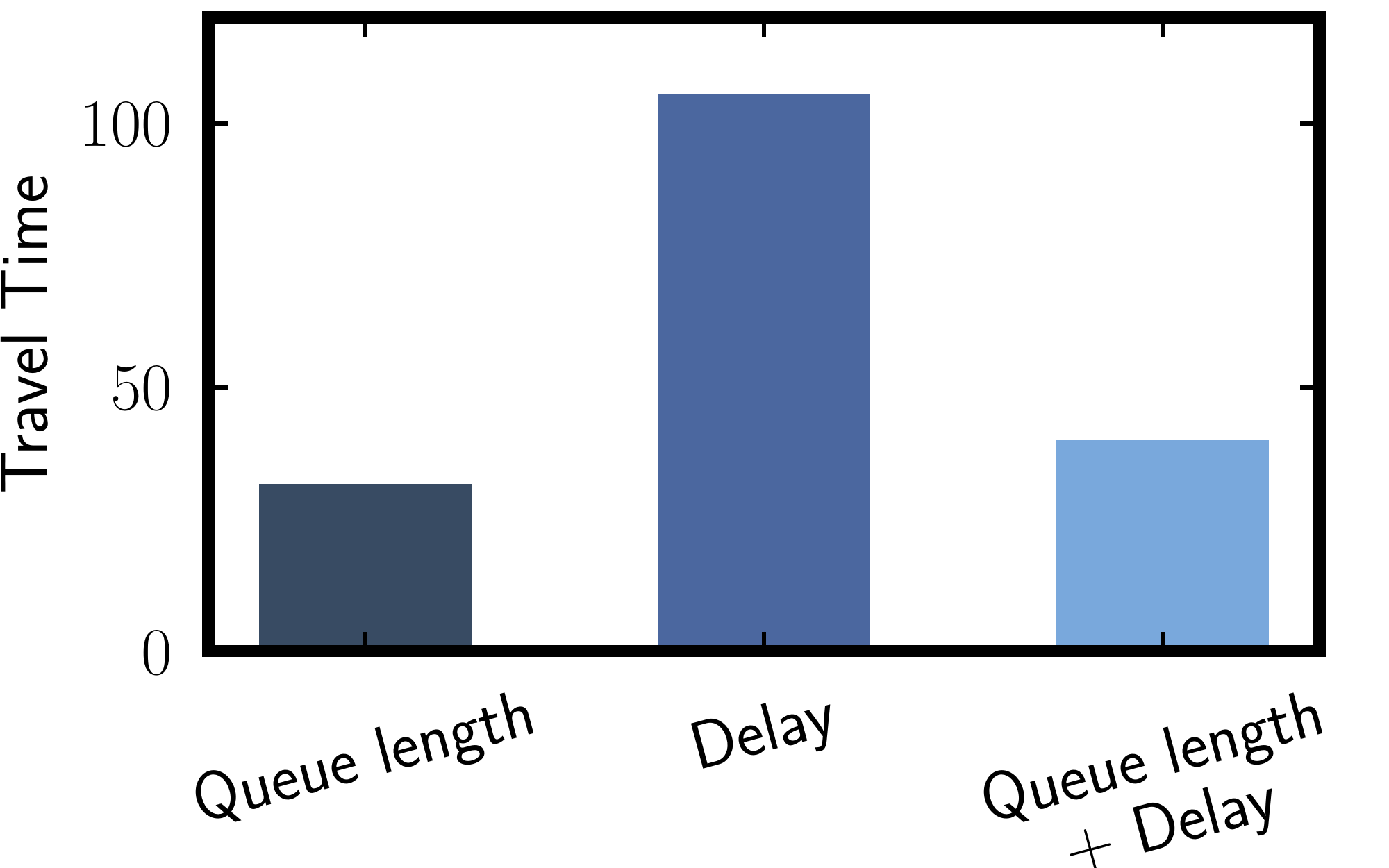}  & 
\hspace{-3mm}\includegraphics[width=0.24\textwidth]{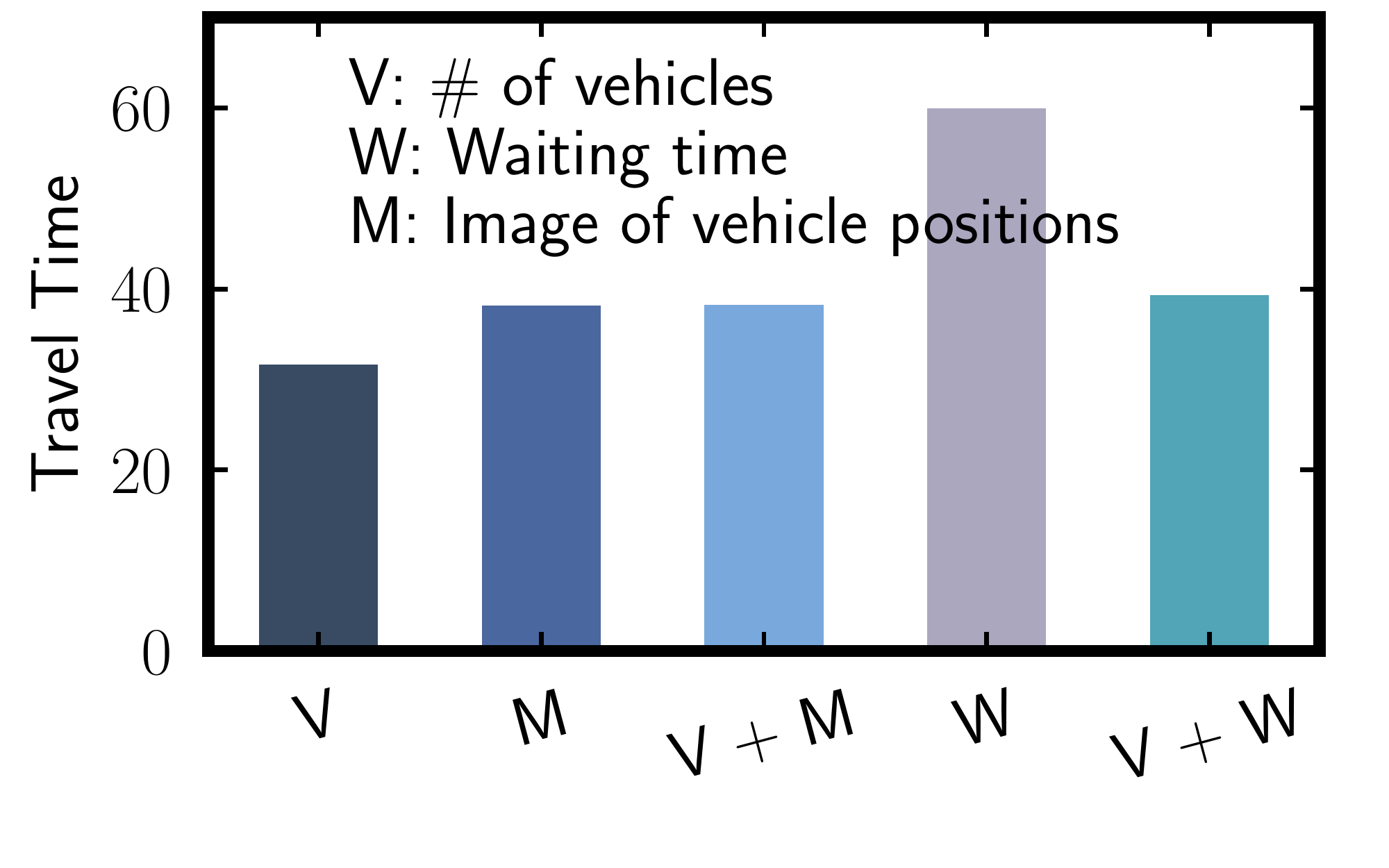} \\
  (a) Different reward used  & (b) Different state used 
\end{tabular}
\caption{Performance with different rewards and states.}
\label{fig:different-state-reward}
\end{figure}

\emph{A key research question for applying RL to traffic signal control is how to define the state and reward.} For the reward, although the ultimate objective is to minimize the travel time of all vehicles, travel time is influenced not only by the traffic lights, but also by other factors like to the free flow speed of a vehicle, thus cannot serve as an effective reward in RL. Therefore, existing studies~\cite{Wier00,VaOl16,wei2018intellilight} typically define reward as a weighted linear combination of several other traffic measures such as queue length, waiting time, number of switches in traffic signal, and total delay. However, there are two concerning issues with these ad-hoc designs. First, there is no guarantee that maximizing the proposed reward is equivalent to optimizing travel time since they are not directly connected in transportation theory. Second, tuning the weights for each reward function component is rather tricky and  minor differences in the weight setting could lead to dramatically different results. This is illustrated in Figure~\ref{fig:different-state-reward}(a): although waiting time and queue length are both correlated with travel time, different weighted combinations of them yield very different results. Unfortunately, there is no principled way to select those weights.
 
In view of this difficulty, in this paper we turn to classic transportation theory to guide us to define the reward. In transportation field, the most classic traffic signal control is based on Webster's delay formula~\cite{webster1966traffic, Roess2011t}. Webster~\cite{webster1966traffic} proposes to optimize travel time under the assumption of uniform traffic arrival rate and to focus only on the most critical intersection volumes (i.e., those that have the highest volume to capacity ratio). Interestingly, critical volume equals to queue length for one time snapshot. Inspired by this connection, we define reward simply as the queue length and we prove that optimizing queue length is the same as optimizing travel time. Experiments demonstrate that this simple reward design consistently outperforms the other complex, ad-hoc reward designs. 

Furthermore, for the state, we observe that there is a trend of using more complicated states in RL-based traffic signal control algorithms in the hope to gain a more comprehensive description of the traffic situation. Specifically, recent studies propose to use images~\cite{VaOl16,wei2018intellilight} to represent the state, which  results in a state representation with thousands or more dimensions. Learning with such a high dimension for state often requires a huge number of training samples, meaning that it takes a long time to train the RL agent. And more importantly, longer learning schedule does not necessarily lead to any significant performance gain. As shown in Figure~\ref{fig:different-state-reward}(b), adding image into state (V+M) actually leads to worse performance than simply using the number of vehicles (V) as the state, as the system may have a harder time extracting useful information from the state representation in the former case. For this reason, we ask in this paper: is there a minimal set of features which are sufficient for the RL algorithm to find the optimal policy for traffic signal control? We show that, if the reward is defined as the queue length, we only need one traffic feature (i.e., the number of vehicles) and the current traffic signal phase to fully describe the system.

To validate our findings, we conduct comprehensive experiments using both synthetic and real datasets. We show that our simplified setting achieves surprisingly good performance and converges faster than state-of-the-art RL methods. We further investigate why RL can perform better than traditional transportation methods by examine three important traits of RL. We believe this study can serve as an important building block for future work in RL for traffic signal control.


In summary, the contributions of this paper are:

\begin{itemize}
\item We propose a simple state and reward design for RL-based traffic signal control. Our reward is simply the queue length, and our state representation only contains two components: number of vehicles on each approaching lane, and the current signal phase. This simplified setting achieves surprisingly good performance on both synthetic data and real data.

\item We draw a connection between our proposed RL method and the classic method in transportation field. We prove that our proposed RL method can achieve the same result as optimal signal timings obtained using classical traffic flow models under steady traffic. 
 
\item We systematically study the key factors that contribute to the success of RL.  Extensive experiments are conducted to show that, under the complex scenarios of real traffic data, our RL method achieves better result than classic transportation methods.
\end{itemize}


\section{Related Work}
\textbf{Classic transportation approaches.} Many traffic signal control approaches have been proposed in the transportation research field. The most common method examines traffic patterns during different time periods of day and creates fixed-time plans correspondingly~\cite{Roess2011t}. However, such simple signal setting cannot adapt to the dynamic traffic volume effectively. Therefore, there have been lots of attempts in transportation engineering to set signal policies according to the traffic dynamics. These methods can be categorized into two groups: actuated methods and adaptive methods.

\textit{Actuated methods}~\cite{mirchandani2001real, fellendorf1994vissim} set the signal according to the vehicle distribution on the road. One typical way is to use thresholds to decide whether the signal should change (e.g., if the queue length on the red signal direction is larger than certain threshold and the queue length on the green signal direction is smaller than another threshold, the signal will change). Obviously the optimal threshold for different traffic volumes is very different, and has to be tuned manually, which is hard to implement in practice. Therefore, these methods are primarily used to accommodate minor fluctuations under relatively steady traffic patterns.

\textit{Adaptive methods} aim at optimizing metrics such as travel time, degree of saturation, and number of stops under the assumption of certain traffic conditions ~\cite{baang1976optimal,silcock1997designing,wong2003lane,haddad2010optimal}. For instance, the widely-deployed signal control systems nowadays, such as SCATS~\cite{scats,scats90,akcelik1981traffic} and SCOOT~\cite{hunt1981scoot,hunt1982scoot,akcelik1981traffic}, choose from manually designed signal plans (signal cycle length, phase split and phase offset) in order to optimize the degree of saturation or congestion. They usually assume the traffic will not change dramatically in the next few signal cycles. Another popular theory to minimize the travel time is proposed by Webster~\cite{webster1966traffic, Roess2011t}. This theory proposes to calculate the minimal cycle length that can satisfy the need of the traffic, and set the phase time proportional to the traffic volume associated with each signal phase. 

\nop{
The aforementioned assumptions set by the transportation methods often do not hold in the real world and therefore prevent these methods from being widely applied. We refer the interested readers to ~\cite{robertson1979traffic,stevanovic2010adaptive} for a more comprehensive survey. Although the traditional transportation methods can not be applied directly in the real world problem, we aim to utilize the theory to guide the design of our RL traffic signal control method.
}

\smallskip
\noindent\textbf{RL approaches.} Recently, reinforcement learning algorithms~\cite{wiering2000multi,mannion2016experimental,Kuyer2008,ElAA13} have shown superior performance in traffic signal control. Typically, these algorithms take the current traffic condition on the road as state, and learn a policy to operate the signals by interacting with the environment. These studies vary in four aspects: RL algorithm, state design, reward design, and application scenario. 
\begin{enumerate}
\item In terms of RL algorithm, existing methods use tabular Q-learning 
\cite{APK03,ElAb10}, deep Q-learning~\cite{wei2018intellilight, liang2018deep}, policy gradient methods~\cite{mousavi2017traffic}, or actor-critic based methods~\cite{casas2017deep}.
\item Different kinds of state features have been used, e.g., queue length~\cite{wei2018intellilight,liLW16}, average delay~\cite{GenR16,drl}, and image features~\cite{gao2017adaptive, liu2017cooperative,drl}.
\item Different choices of reward includ average delay~\cite{arel2010reinforcement,drl}, 
average travel time~\cite{liu2017cooperative}, 
and queue length~\cite{wei2018intellilight,liLW16}.  
\item Different application scenarios are covered in previous studies, including single intersection control~\cite{wei2018intellilight}, and multi-intersection control~\cite{ElAA13,ALUK10,da2006adaptive}.
\end{enumerate}

To date, these methods mainly focus on designing complex state or reward representations to improve the empirical performance, without any theoretical support. But complex designs typically require a large quantity of training samples and longer training process, which could lead to potential traffic jam, and large uncertainty in the learned model.

In this paper, we will illustrate how a simpler design of state and reward function could actually help RL algorithms find better policies, by connecting our design with existing transportation theory.
\nop{Although we investigate the problem in a simple scenario, the finds in the paper are still essential when migrating to a more complex multi-intersection scenario and more complex RL model design.}

\smallskip
\noindent \textbf{Interpretable RL.} Recently, there are studies on interpreting the ``black box'' RL models. Most of them use self-interpretable models when learning the policies, i.e., to mimic the trained model using interpretable models~\cite{VMSK18,HeUR2017}. This line of work serves a different goal from our work. Rather than providing a general interpretation of RL model, we aim to draw a connection with classic transportation theory and to examine the key factors which lead to the superior performance of RL.


\section{Problem Definition}
\subsection{Preliminary}
In this paper, we investigate the traffic signal control systems. The following definitions in a four-way intersection will be used throughout this paper.
\begin{itemize}
    \item \textbf{Entering direction:} The entering directions can be categorized as West, East, North and South (`W', `E', `N', `S' for short).
    \item \textbf{Signal Phase:} A signal light mediating the traffic includes three types: Left, Through, and Right (`L', `T', and `R' for short). A complete  signal phase is composed of two green signal lights and other red lights. The signal phase is represented as the format of $A_1B_1$-$A_2B_2$,  where $A_i$ denotes one entering direction, and $B_i$ is one type of signal light. For instance, \gWEStraight means green lights on west and east through traffic and red lights on others.
\end{itemize}

\subsection{RL Environment}
\label{sec:rl_env}

The RL environment describes the current situation in an intersection, e.g., current signal phase and positions of vehicles. A RL agent will observe the environment and represent it as a numerical state representation $s_t$, at each timestamp $t$. 

A pre-defined signal phase order is given as ($p^0$, $p^1$, $p^2$, ..., $p^{K-1})$.
Each timestamp, the agent will make a decision on whether to change the signal to next phase ($a_t$ = 1) or keep the current signal ($a_t$ = 0). Note that a pre-defined phase order is the common practice in transportation engineering~\cite{Roess2011t,urbanik2015signal}, as it aligns with drivers' expectation and can avoid safety issues. 

The action $a_t$ will be executed at the intersection and the intersection will come to a new state $s_{t+1}$. Further, a reward $r$ is obtained from the environment, which can be described by a function $R(s_t, a_t)$ of state $s_t$ and action $a_t$. Then, the traffic signal control can be formulated as the classic RL problem:

\begin{problem}
Given the state observations set $\mathcal{S}$, action set $\mathcal{A}$, the reward function $R(s, a)$. The problem is to learn a policy $\pi(a|s)$, which determines the best action $a$ given the state $s$, so that the following expected discounted return is maximized.
\begin{equation}
    G_t = R_{t+1} + \gamma R_{t+2} + \gamma^2 R_{t+3} + ... = \sum_{b=0}^{\infty} \gamma^b R_{t+b+1}.
    \label{eq:return}
\end{equation}
\end{problem}
Specifically, the state, action and reward are defined as below.
\begin{itemize}
\item {\bf State}: The state includes the number of vehicles $v_{j,t}$ on each lane $j$ and the current signal phase $p_t$.
\item {\bf Action}: When signal changes, $a_t=1$, otherwise $a_t=0$.
\item {\bf Reward}: The reward is defined as the summation of queue length over all lanes:
\begin{equation}
    R_t = - \sum_{j=1}^{M} q_{t, j}.
    \label{eq:reward-def}
\end{equation}
\end{itemize}

Table~\ref{tab:notation} summarizes the key notation used throughout this paper. 

\begin{table}[htb]
\centering
  \caption{Notation}
  \label{tab:notation}
  \begin{tabular}{ll}
    \toprule
    Notation&Meaning\\
    \midrule
    $s$      & State                       \\
    $a$        & Action                      \\
    $R$        & Reward                      \\
    $q_j $ & Queue length on the lane $j$ \\
	$v_j$ & Number of vehicles on the lane $j$ \\
	$p$ & Signal phase\\
	$K$ & Number of signal phase \\
	$M$ & Number of lanes \\
	$N$ & Number of vehicles in system \\
	 \bottomrule
\end{tabular}
\end{table}

\subsection{Objective}
We set the goals of this paper as follows:
\begin{itemize}
\item Find an RL algorithm which to solve the above problem.
\item Connect the RL algorithm with classic transportation theory, and prove the optimality of the RL algorithm with our concise state and reward definition.
\item Analyze the traits that make RL outperform other methods.
\end{itemize}


\section{Method}
\label{sec:method}
Solving traffic signal control problem via RL has attracted lots of attention in recent years~\cite{Wier00,wei2018intellilight}. Among these studies, much efforts have been made to the design of reward function and state representation, without understanding why RL performs well in practice. As a consequence, some rewards and states may be redundant and have no contribution to performance improvement. 
In this section, we first propose a simple but effective design of RL agent to tackle the traffic signal control problem. Then, we illustrate why this simple reward and state design may enable RL to reach the optimal solution. Finally, we describe three traits that make RL outperform the other methods. 

\subsection{Sketch of Our RL Method}
\begin{figure*}
\centering
  \includegraphics[width=0.95\textwidth]{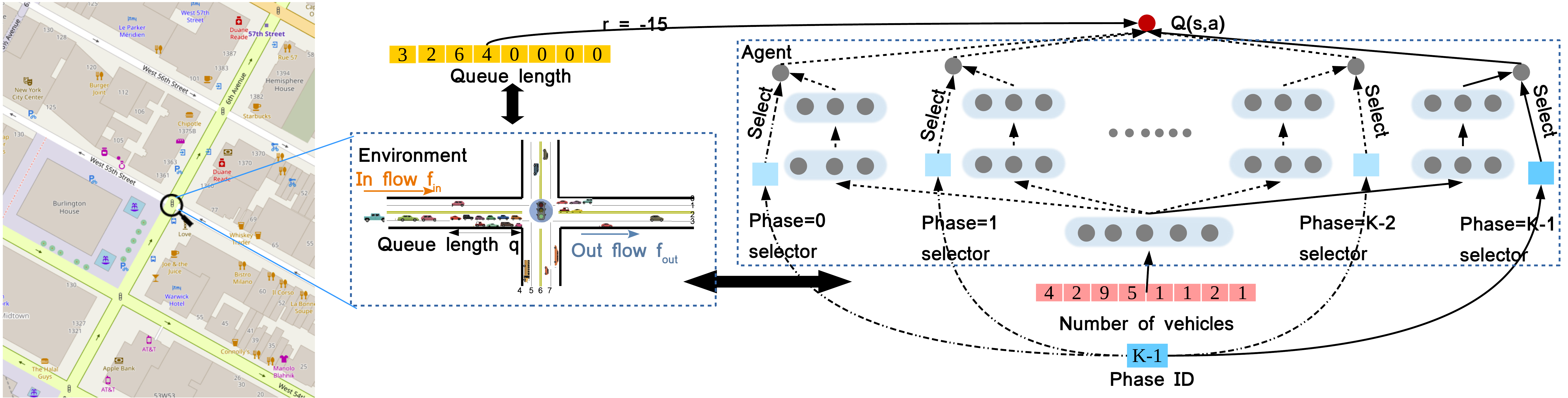}
     \caption{Overview of our traffic signal control system.}
    \label{fig:network}
\end{figure*} 
We name our method as \textbf{\underline{L}}ight-\textbf{\underline{I}}ntelligh\textbf{\underline{T}} (\deeplightQueue), which adopts the design of state and reward in Section~\ref{sec:rl_env}.
We employ the Deep Q-Network proposed by~\cite{wei2018intellilight} to seek the action that achieves the maximum long-term reward defined by Bellman Equation~\cite{RiAn98}:
\begin{equation}
Q(s_t,a_t)=R(s_t, a_t)+\gamma \max Q(s_{t+1}, a_{t+1}).
\label{eq:bellman}
\end{equation}

As illustrated in Figure~\ref{fig:network}, all features, i.e., the vehicle number $v_j$ and phase indicator $p$ are concatenated first to generate the feature embedding via the fully-connected layers. $K$ distinct paths are enabled towards the final reward estimation by the phase selector structure. This structure has been proven to be effective~\cite{wei2018intellilight} in enabling different state-action-reward mapping function learning under different phases.

\subsection{Classic Transportation Theory}
\label{sec:trans-theory}
Before looking into RL-based traffic signal control algorithms, we first discuss how to solve this problem from the transportation research perspective. Typically transportation approaches aim to minimize the total travel time under a certain traffic condition. The problem can be mathematically expressed as:

\begin{equation*}
\begin{aligned}
& {\text{minimize}}
& & \sum_{j=1}^{M}{T_j} \\
& \text{subject to}
& & g^{min}<=g_k<=g^{max} \\
&&& \frac{g_j}{C} >= \frac{f_{in, j}}{u_{sat, j}} \; j = 1, 2, 3, ..., M.
\end{aligned}
\end{equation*}
Here, $T_j$ is the total travel time for vehicles approaching from lane $j$, $g_{k}$ is the green time allocated to phase $k$ [sec], $g_{j}$ is the green time allocated to lane $j$ [sec], $C$ is the cycle length at the intersection, $f_{in, j}$ is car arrival flow in lane $j$ [vehicles/hour], and $u_{sat, j}$ is the saturation flow in lane $j$ [vehicles/hour].

Note that vehicles from an approaching lane $j$ can be served by multiple phases, and a phase $k$ can serve multiple lanes, so one should not confuse $g_j$ with $g_k$. The first constraint set lower and upper bounds for each green phase and the second constraint ensures that each lane gets appropriate green phase length. Assuming uniform vehicle arrivals and no left-over queues for all lanes at the intersection, the \textit{Webster' delay formula}~\cite{webster1966traffic} can be used. The total travel time at an intersection can be estimated as total free-flow travel time (a constant for a lane based on the intersection geometry and free-flow travel speed) plus total delay, which is expressed as:

\begin{equation*}
d_j = \sum_{j=1}^{M}{\frac{f_{in, j}}{1-f_{in, j}/u_{sat, j}}\times \lambda_{j}}
\end{equation*}
where $d_j$ is the total delay for lane $j$ and $\lambda_{j}$ is the time interval between the green phases that serve lane $j$ in two consecutive cycles.

There exists a closed-form solution for the optimal signal cycle length in this case, which is expressed as follows
:
\begin{equation}
C_{des} = \frac{K\times t_L}{1-V_c/\frac{3600}{h}},
\label{eq:transportation-theory}
\end{equation}
where $K$ is the number of phases; $t_L$ is the total loss time per phase; $h$ is the saturation headway [seconds/vehicle], defined as the smallest time interval between successive vehicles passing a point; $V_c$ is the sum of critical lane volumes from the most constraining movement served by each signal phase. 

The solution is commonly used in practice to calculate the desired cycle length at intersections, and then the time allocated to each phase is split according to the critical traffic volume in each lane group. However, these calculations are commonly based on several sets of typical traffic demands (e.g., AM peak, mid-day PM peak, etc.), and therefore can not adapt to dynamic traffic well. Even if the traffic demands are sampled frequently, the usage of a fixed equation may not fit all intersections appropriately. 

\subsection{Connecting RL with Transportation Theory}
\label{sec-method}

In this section, we aim to build connections between our RL algorithm and transportation methods by illustrating that, in a simplified transportation system,
(1) using queue length as reward function in RL is equivalent to optimizing travel time as in the transportation methods, and (2) the number of vehicles on the lane $j$, $v_j$, and phase $p$ fully describe the system dynamics.

\subsubsection{Queue length is equivalent to travel time} 

To understand our reward function, suppose there is an intersection with $M$ lanes in total considering all entering directions. We consider a set of vehicles with size $N$. Suppose the first vehicle arrives at the intersection at time $t=0$, and the last vehicle passes the intersection at $\tau$. In the following, we only focus on the vehicles in this set when calculating the queue length and travel time. But note that our calculation is general and applies to arbitrary number of vehicles. 

First, by setting $\gamma=1$ and restricting our attention in the time interval $[0,\tau]$, we can rewrite our RL objective Eq.~\eqref{eq:return} as follows:

\begin{equation*}
\begin{aligned}
& \underset{\pi}{\text{maximize}}
& & \sum_{t=1}^{\tau} R_t(s, a). \\
\end{aligned}
\end{equation*}

Now, by substituting the queue length as the reward, and dividing the objective by a constant $\tau$, we reach the following formulation:
\begin{equation*}
\begin{aligned}
& \underset{\pi}{\text{minimize}}
& & \frac{1}{\tau}  \sum_{t=1}^{\tau} \sum_{j=1}^M q_{t,j} \doteq \bar{q}, \\
\end{aligned}
\end{equation*}
where $\bar{q}$ is the average queue length of all the lanes during the time interval $[0,\tau]$.

Next, let us define a \emph{waiting event} as the event that a vehicle waits at the intersection for a unit time. Then, the value of $e_{t, i}$ for vehicle $i$ at timestamp $t$ is defined as ~\eqref{eq:event}. 
\begin{equation}
    e_{t, i} = 
    \begin{cases}
        0 & \text{vehicle } i \text{ is not waiting at timestamp } t \\
        1 & \text{vehicle } i \text{ is waiting at timestamp } t
    \end{cases}
\label{eq:event}
\end{equation}
Note that, when vehicle $i$ is not waiting, there could be three different cases:
\begin{itemize}
    \item Vehicle $i$ is moving on the approaching lane.
    \item Vehicle $i$ has not arrived on the approaching lane.
    \item Vehicle $i$ has left the approaching lane.
\end{itemize}

Next, it is easy to see that, the number of waiting events generated at time $t$, $e_t$, can be computed as the sum of queue length of all lanes at this timestamp:
\begin{equation}
    e_{t} = \sum_{j=1}^M q_{t, j}.
\end{equation}
Then, the total waiting events generated by all the $N$ vehicles during the interval $[0, \tau]$ can be expressed as 
\begin{equation}
W = \sum_{t=1}^{\tau} e_t = \sum_{t=1}^{\tau} \sum_{j=1}^M q_{t, j} = \tau \times \bar{q}.
\label{eq:sum-queue}
\end{equation}

In the meantime, the delay that vehicle $i$ experiences during its trip across the intersection, which represents the additional travel time over the time the vehicle needs to physically move through the intersection (i.e., if it is the only vehicle in the system and there is no traffic signal), can be calculated as 
\begin{equation}
    D_i = \sum_{t = 1}^{\tau} e_{t, i}.
\end{equation}
Then, it is evident that the average travel time of the vehicles can be calculated as follows:
\begin{equation}
    \bar{T} =  \frac{1}{N} \sum_{i = 1}^{N} (D_i + l/\mu) = \frac{1}{N} \sum_{i = 1}^{N} (\sum_{t = 1}^{\tau} e_{t, i} + l/\mu) = \frac{W}{N} + l/\mu,
    \label{eq:travel-time}
\end{equation}
where $l$ is the length of the road, and $\mu$ is the free flow speed of a vehicle.
Finally, substituting \eqref{eq:sum-queue} into \eqref{eq:travel-time}, we have
\begin{equation}
\bar{T} = \frac{\tau \times \Bar{q}}{N} + l/\mu.
\label{eq:tt-equal-q}
\end{equation}
From Eq.\eqref{eq:tt-equal-q}, we can see that $\bar{T}$ is proportional to the average queue length $\bar{q}$ during the time interval $[0, \tau]$. In other words, by minimizing the average queue length, the RL agent is also minimizing the average travel time of the vehicles! Therefore, in this paper, we use queue length as the reward function. We will further show the effectiveness of our reward function in the experiments.

\subsubsection{Number of vehicles on the lane $j$ ($v_j$) and the phase $p$ fully describe the system dynamics} To further understand why we use $v_j$ and $p$ as the only state features in the Deep Q-Network, we now show that the dynamics of the system can be fully determined by these two variables, when the traffic arrive at the intersection uniformly. 

Specifically, let $p^k$ represents the $k$th phase in the pre-defined phase sequence $(p^0, p^1, p^2, ..., p^{K-1})$. Then, given the current phase at time $t$, $p_t = p^k$, the system transition function of the lane is:
\begin{equation}
v_{t+1, j} = v_{t, j} + f_{in, j} - f_{out, j}\times c_{t, j},
\end{equation}
where 
\begin{equation}
    c_{t, j} = 
    \begin{cases}
        0 & \text{lane } j \text{ is on red light at timestamp } t \\
        1 & \text{lane } j \text{ is on green light at timestamp } t
    \end{cases}
\label{eq:ct}
\end{equation}
and
\begin{equation}
    p_{t+1} = 
    \begin{cases}
        p^k & a_t = 0 \\
        p^{(k+1) \text{ mod } K} & a_t = 1
    \end{cases}
\label{eq:pt}
\end{equation}
where mod stands for the modulo operation.

Note that while we assume $f_{in, j}$ and $f_{out, j}$ are constant, they may not be known to the agent. However, $f_{in, j}$ can be easily estimated by the agent from $v_j$ at any timestamp when the red light is on: $f_{in, j} = v_{t+1, j} - v_{t, j}, \forall t: c_{t, j} = 0$; and $f_{out, j} $ can then be estimated when the green light is on: $f_{out, j} = v_{t, j} - v_{t+1, j} + f_{in, j}, \forall t: c_{t, j} = 1$.
Therefore, in theory, using $v_j$ and $p$ as the only features, it is possible for the agent to learn the optimal policy for the system. Although this estimation of system dynamics relies on the assumption that the in flow and out flow are relatively steady, we note that real traffic can often be divided into segments with relatively steady traffic flow in which similar system dynamics are applicable. Therefore, our simple state design can be a good approximation when complex traffic scenario is considered.

\subsection{Analysis of Traits of RL Approach}

In the previous section, we have demonstrated the optimality of our method under simple and steady traffic. However, in most cases where the traffic pattern is complicated, methods from traditional transportation engineering cannot guarantee an optimal solution, and even fail dramatically. For instance, the conventional transportation engineering methods may not adjust well to the rapid change of traffic volume during peak hour. Meanwhile, RL has shown its superiority in such cases. In the next, we discuss three traits that make RL methods outperform the traditional methods.

\subsubsection{Online learning}
RL algorithms can get feedback from environment and update their models accordingly in an online fashion. If an action leads to inferior performance, the online model will learn from this mistake after model update.

\subsubsection{Sampling guidance}
Essentially, in this problem, the algorithm aims to search for the best policy for the current situation. RL will always choose an action based on experience learned from past trials. Hence, it will form a path with relatively high reward. In contrast, a random search strategy without guidance will not utilize the previous good experience. This will lead to slow convergence and low reward (i.e., severe traffic jams in the traffic signal control problem) during the search progress. 

\subsubsection{Forecast}
The Bellman equation Eq.~\eqref{eq:bellman} enables RL to predict future reward implicitly through the Q-function. This could help the agent take actions which might not yield the highest instant rewards (as most methods in control theory do), but obtain higher rewards in the long run.


\section{Experiments}

\subsection{Experiment Setting}
Following the tradition of the traffic signal control study~\cite{wei2018intellilight}, we conduct experiments in a simulation platform SUMO (Simulation of Urban MObility)\footnote{\url{http://sumo.dlr.de/index.html}}. After the traffic data being fed into the simulator, a vehicle moves towards its destination according to the setting of the environment. The simulator provides the state to the signal control method and executes the traffic signal actions from the control method. Following the tradition, each green signal is followed by a three-second yellow signal and two-second all red time. 

In a traffic dataset, each vehicle is described as $(o, t, d)$, where $o$ is  origin location, $t$ is  time, and $d$ is  destination location. Locations $o$ and $d$ are both locations on the road network. Traffic data is taken as input for simulator.

In a multi-intersection network setting, we use the real road network to define the network in simulator. For a single intersection, unless otherwise specified, the road network is set to be a four-way intersection, with four 300-meter long road segments.

\subsection{Evaluation Metrics and Datasets}

Following existing studies, we use `travel time' to evaluate the performance, which calculates that the average travel time the vehicles spent on approaching lanes. This is the most frequently used measure for traffic signal performance in transportation field. Other measures indicate similar results and are not shown here due to space limit. Additionally, we will compare some learning methods using the convergence speed, which is the average number of iterations a learning method takes to reach a stable travel time.

\subsubsection{Real-world data} The real-world dataset consists of two private traffic datasets from \jn and \hz, and one public dataset from Los Angeles in the United States.\footnote{We omit the cities' name here to preserve data privacy.}
\begin{itemize}
    \item \textbf{\jn}: This dataset is captured by the surveillance cameras near five road intersections in a city of China. Computer vision methods have been used to extract the the time, location and vehicle information. We reproduce the trajectories of vehicles from these records by putting them into the simulator at the timestamp that they are captured. 
    
    \item \textbf{\hz}: This dataset is collected from the loop sensors in a city of China. One data record is generated in usually 2-3 minutes. Each record contains the traffic volume count in different approaches.
    
    \item \textbf{Los Angeles}: This is a public dataset\footnote{https://ops.fhwa.dot.gov/trafficanalysistools/ngsim.htm} collected from Lankershim Boulevard, Los Angeles on June 16, 2005. This dataset covers four successive intersections  (three are four-way intersections and one is three-way intersection).
\end{itemize}

\subsubsection{Synthetic data} 
Uniform traffic is also generated here with an attempt to justify that our proposed RL model can approach the theoretical optimality under the assumption of ideal traffic environment. Specifically, we generate one typical uniform traffic with the traffic volume as 550 vehicles/lane/hour.

\begin{table*}[ht]
\centering
\caption{Performance on real-world traffic from \jn, \hz and Los Angeles. The reported value is the average travel time of vehicles in seconds at each intersection. The lower the better.}
\label{tab:real_performance}

\begin{tabular}{l|lllll|lll|llll}
\toprule
\multirow{2}{*}{\diagbox[width=7em,trim=l]{Model}{Intersection}} & \multicolumn{5}{c|}{\jn} & \multicolumn{3}{c|}{\hz} & \multicolumn{4}{c}{Los Angeles} \\
 & \multicolumn{1}{c}{1} & \multicolumn{1}{c}{2} & \multicolumn{1}{c}{3} & \multicolumn{1}{c}{4} & \multicolumn{1}{c|}{5} & \multicolumn{1}{c}{1} & \multicolumn{1}{c}{2} & \multicolumn{1}{c|}{3} &
 \multicolumn{1}{c}{1} & \multicolumn{1}{c}{2} & \multicolumn{1}{c}{3} &
 \multicolumn{1}{c}{4}
 \\\midrule
\fixedtime~\cite{Miller1963} & 121.70 & 123.37 & 128.98 & 113.39 & 37.66 & 47.15 & 33.86 & 79.91 & 350.18 & 345.85 & 376.11 & 237.57 \\
\fixedtimeHistory~\cite{Roess2011t} & 83.19 & 62.23 & 52.27 & 56.07 & 39.56 & 41.84 & 46.63 & 64.79 & 248.44 & 118.72 & 135.16 & 130.44 \\
\sotl~\cite{sotl} & 82.14 & 67.18 & 43.51 & 60.54 & 66.73 & 62.10 & 56.85 & 50.40 &  133.84 & 96.15 & 78.96 & 99.63 \\
\midrule
\pg~\cite{mousavi2017traffic} &207.69&128.77&155.32&145.37&51.92
  & 48.74 & 47.16 & 106.74 & 431.56 & 150.63 & 421.71 & 344.12 \\
\nips~\cite{drl} & 119.86 & 113.92 & 136.49 & 117.78 & 89.43 & 138.60 & 172.01 & 151.65& 187.61 & 188.84 & 174.12 & 259.73 \\
\deeplight~\cite{wei2018intellilight} & 122.09 & 81.39 & 58.33 & 56.72 & 36.24 & 42.80 & 36.19 & 51.01 & 90.38 & 90.46 & 85.20 & 90.52 \\
\midrule
\deeplightQueue & \textbf{73.61} & \textbf{57.68} & \textbf{31.21} & \textbf{47.84} & \textbf{31.85} & \textbf{32.51} & \textbf{29.76} & \textbf{44.16} & \textbf{42.16} & \textbf{61.94} & \textbf{68.73} & \textbf{76.41}
\\\bottomrule
\end{tabular}
\end{table*}

\subsection{Compared Methods}
\label{sec-compared-methods}
To evaluate the effectiveness of our model, we compare the performance of the following methods: 

\begin{itemize}

\item \fixedtime: Fixed-time control~\cite{Miller1963} uses a pre-determined cycle and phase time plan (determined by peak traffic) and is widely used in stable traffic flow scenarios. Here, we allocate 30 seconds to each signal phase and 5 seconds yellow light after each phase.
\item \fixedtimeHistory: \fixedtimeHistory~\cite{Roess2011t} calculates the cycle length of the traffic signal according to the traffic volume by Eq.~\eqref{eq:transportation-theory}. As illustrated in Section~\ref{sec:trans-theory}, the time split for each phase is determined by the critical volume ratio of traffic corresponding to each phase.

\item \nop{\textbf{Self-Organizing Traffic Light Control}} \sotl: Self-Organizing Traffic Light Control~\cite{sotl} is an adaptive method which controls the traffic signal with a hand-tuned threshold of the number of waiting cars. The traffic signal will change when the number of waiting cars exceeds a hand-tuned threshold.
\item \pg: Policy gradient \cite{mousavi2017traffic} is another way of solving the traffic signal control problem using RL. It uses the image describing positions of vehicles as state representation, and the difference of cumulative delay before and after action as reward. Different from the value-based RL methods, this method parameterizes and optimizes the policy directly. 

\item \nop{\textbf{Deep Reinforcement Learning for Traffic Light Control}} \nips: This method~\cite{drl} is a deep RL method that employs the DQN framework for traffic signal control. It uses an image describing the positions of vehicles on the road as state and combines several measures into the reward function, such as flickering indicator, emergency times, jam times, delay, and waiting time. 
\item \deeplight: This method~\cite{wei2018intellilight} is also a deep RL method which uses richer representations of the traffic information in the state and reward function. With a better designed network architecture in the DQN framework, it achieves the state-of-the-art performance for traffic signal control.

\item \deeplightQueue: This is our proposed method.
\end{itemize}

\subsection{Overall Performance}

\subsubsection{Real-world traffic}
The overall performance of different methods in real-word traffic is shown in Table~\ref{tab:real_performance}. We can observe that our proposed model \deeplightQueue consistently outperforms all the other transportation and RL methods on all three datasets. Specifically, some RL methods like \pg and \nips perform even worse than the classic transportation methods, such as \formula and \sotl, because of their simple model design and the large training data needed. \deeplight performs better, but still lags behind \deeplightQueue due to its use of multiple complicated components in the state and reward function. 

\subsubsection{Uniform traffic flow}

\begin{figure}[ht]
\centering
 \includegraphics[width=0.45\textwidth]{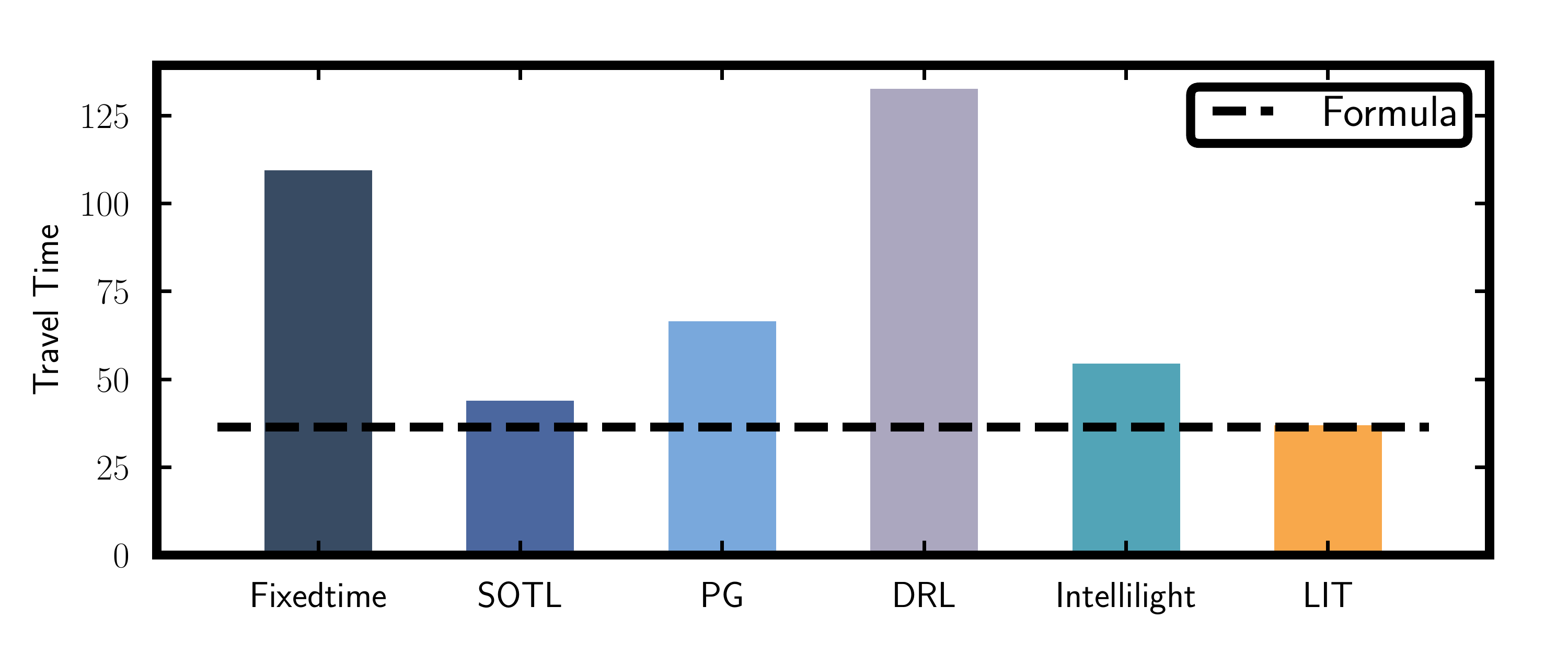}
\caption{Performance on uniform traffic with volume = 550 vehicles/lane/hour. \deeplightQueue outperforms other baseline methods and approaches the theoretical optimality in the uniform traffic (\formula).}
\label{fig:uniform_cmp}
\end{figure}

As elaborated in Section~\ref{sec-method}, our definition of state and reward is sufficient for a RL algorithm to learn the optimal policy under uniform traffic flow. Here, we validate it by comparing \deeplightQueue with \formula, which is guaranteed to achieve the theoretical optimality in the simplified transportation system. As shown in Figure~\ref{fig:uniform_cmp}, 
\deeplightQueue outperforms all the other baseline methods and achieves almost identical performance with \formula.

\begin{figure}[h]
  \centering
  \includegraphics[width=0.45\textwidth]{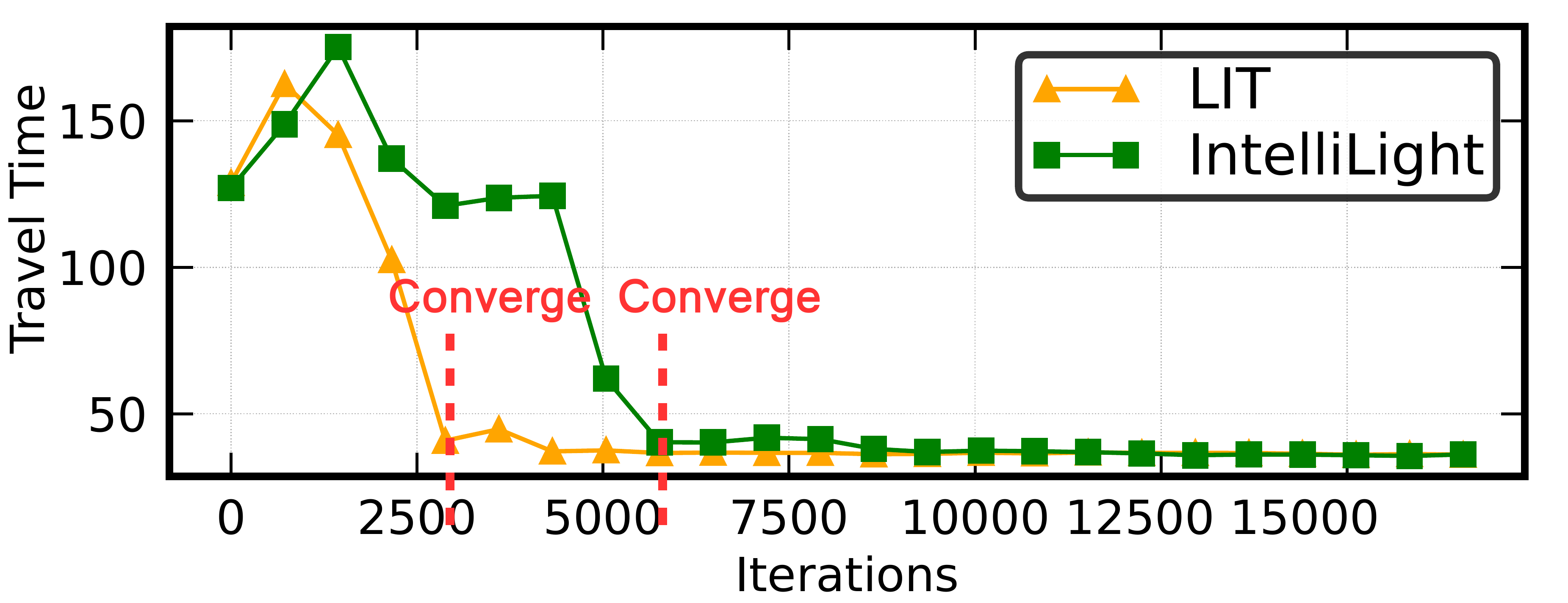}
     \caption{Convergence comparison between \deeplight and \deeplightQueue in uniform traffic. \deeplightQueue converges after 2,880 iterations, while \deeplight converges after about 5,760 iterations.}
    \label{fig:converge}
\end{figure}

Besides the improvement in performance, \deeplightQueue also greatly reduce the training time on account of succinct state representation and lightweight network. As depicted in Figure~\ref{fig:converge}, the convergence speed of \deeplightQueue in uniform traffic is twice faster than that of \deeplight.

\subsection{Performance in Multi-intersection Network}

In this section, we further explore the traffic signal control in more complex scenarios including different multi-intersection structures with an attempt to justify the universality of our model.
Note that,  the traffic signal control in multi-intersection is still an open and challenging problem in the literature. Some prior work studies various coordination techniques in multi-intersection setting, but few of them have realized explicit coordination. And regardless of the type of coordination, individual control is always the foundation. Further, Girault et.al.~\cite{girault2016exploratory} recently argued that coordination between intersections may not be a necessity in the road network setting. 
As a result, we just focus on the individual control in the multi-intersection scenarios.

\begin{figure}[ht]
\centering
 \includegraphics[width=0.45\textwidth]{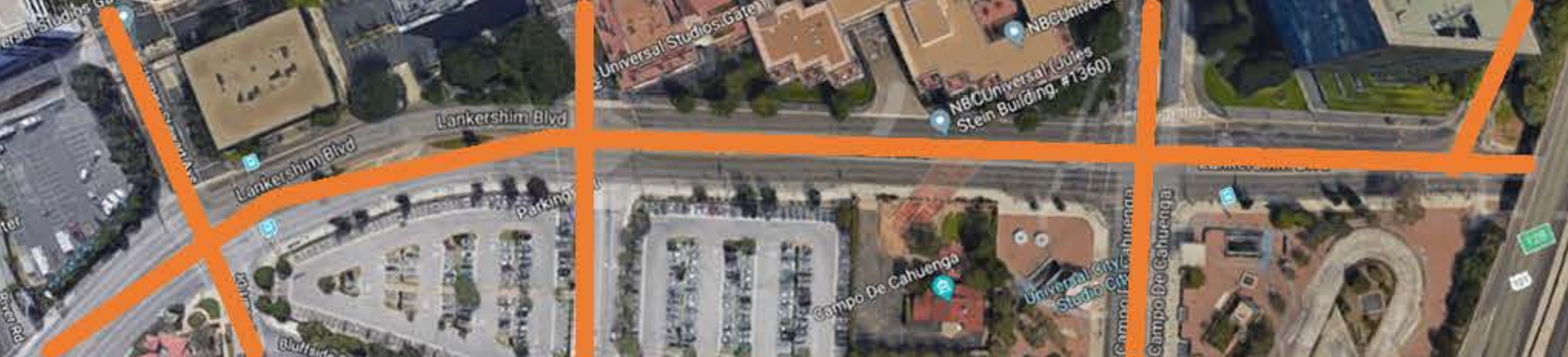}
\caption{The satelite of $1\times4$ grid of intersection in Los Angeles. One intersections only has three entering way.}
\label{fig:la_map}
\end{figure}

In this experiment, we use three types of road network, including $3\times4$, $4\times4$, and $1\times4$ grids, with the traffic data collected from \jn, \hz, and Los Angeles, respectively. The satellite image of the four intersections in Los Angeles is shown in Figure~\ref{fig:la_map}.  For this experiment, \pg is not included as it has the worst performance and slowest convergence speed in the single intersection setting. As depicted in Table \ref{tab:hangzhou_4intersections}, \deeplightQueue again outperforms other baselines. Notably, other RL methods sometimes have even worse performance than classic transportation methods, which suggests that the definition of reward and state plays an critical role in the effectiveness of RL-based methods, especially for complex (multi-intersection) scenario.

\begin{table}[h]
\centering
\caption{Performance in multi-intersection environment. The reported values are the average travel time of vehicles. The lower the better.}
\label{tab:hangzhou_4intersections}
\begin{tabular}{lccc}
\toprule
Model&\jn&\hz & Los Angeles\\\midrule
\fixedtime~\cite{Miller1963} &804.63&572.60 & 682.50\\
\fixedtimeHistory~\cite{Roess2011t}&440.13&419.13 & 831.34\\
\sotl~\cite{sotl} &1360.24&834.66 &1710.86\\
\midrule
\nips~\cite{drl}&1371.45&1155.64 & 1630.0\\
\deeplight~\cite{wei2018intellilight}&366.26&433.89 & 375.21\\
\midrule
\deeplightQueue&\bf 274.83& \bf 398.92 & \bf 126.63\\\bottomrule
\end{tabular}
\end{table}

\subsection{Case study}
\label{sec:case_study}
\begin{figure}
\begin{tabular}{c}
    \includegraphics[width=0.35\textwidth]{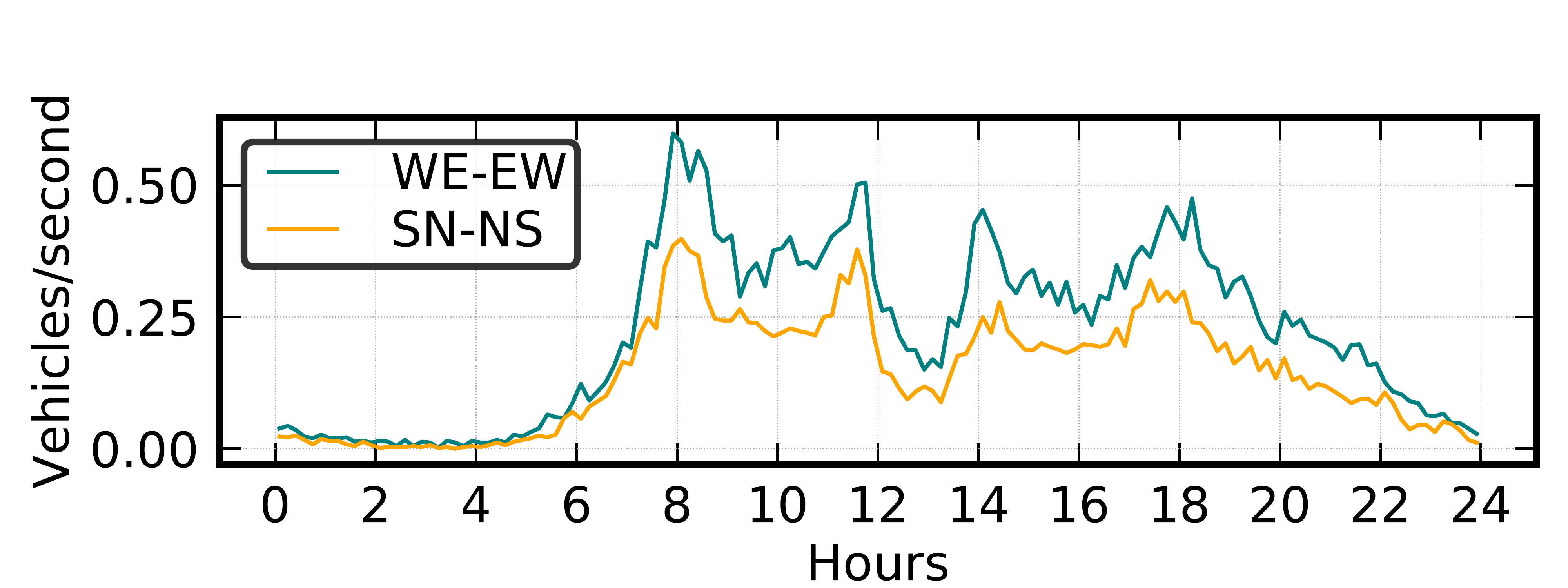}  \\ 
  \includegraphics[width=0.35\textwidth]{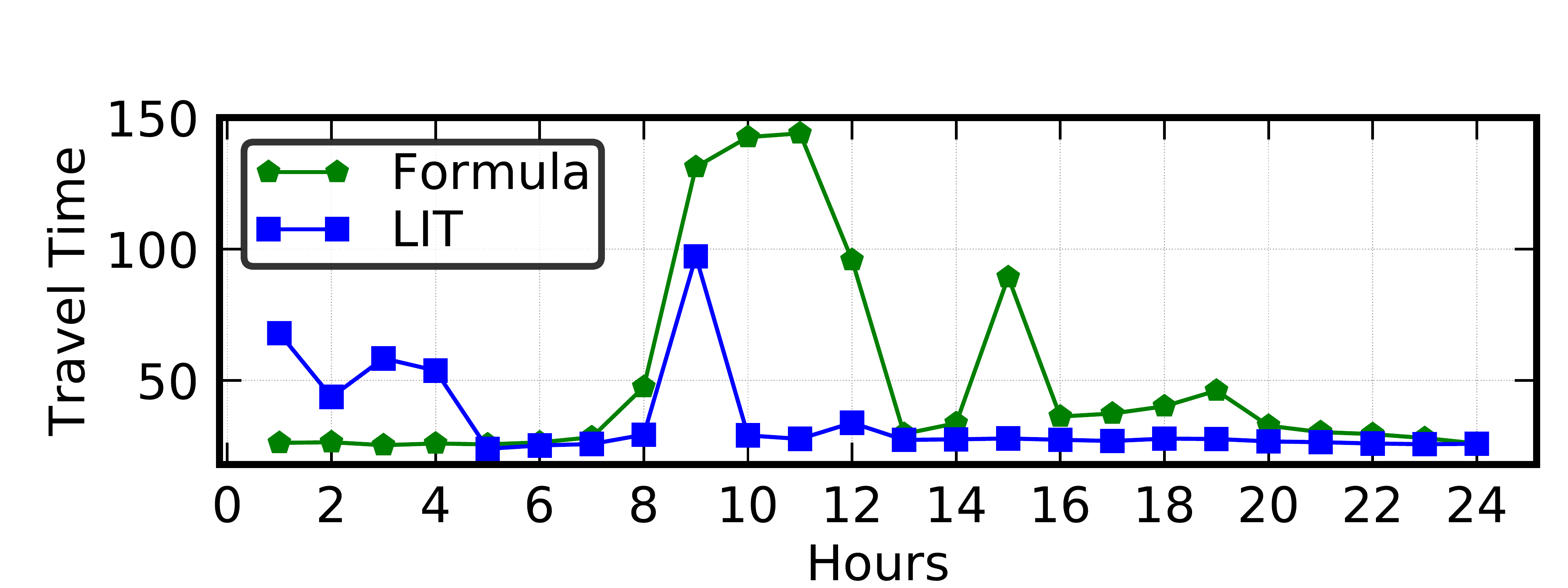} \\
  \includegraphics[width=0.35\textwidth]{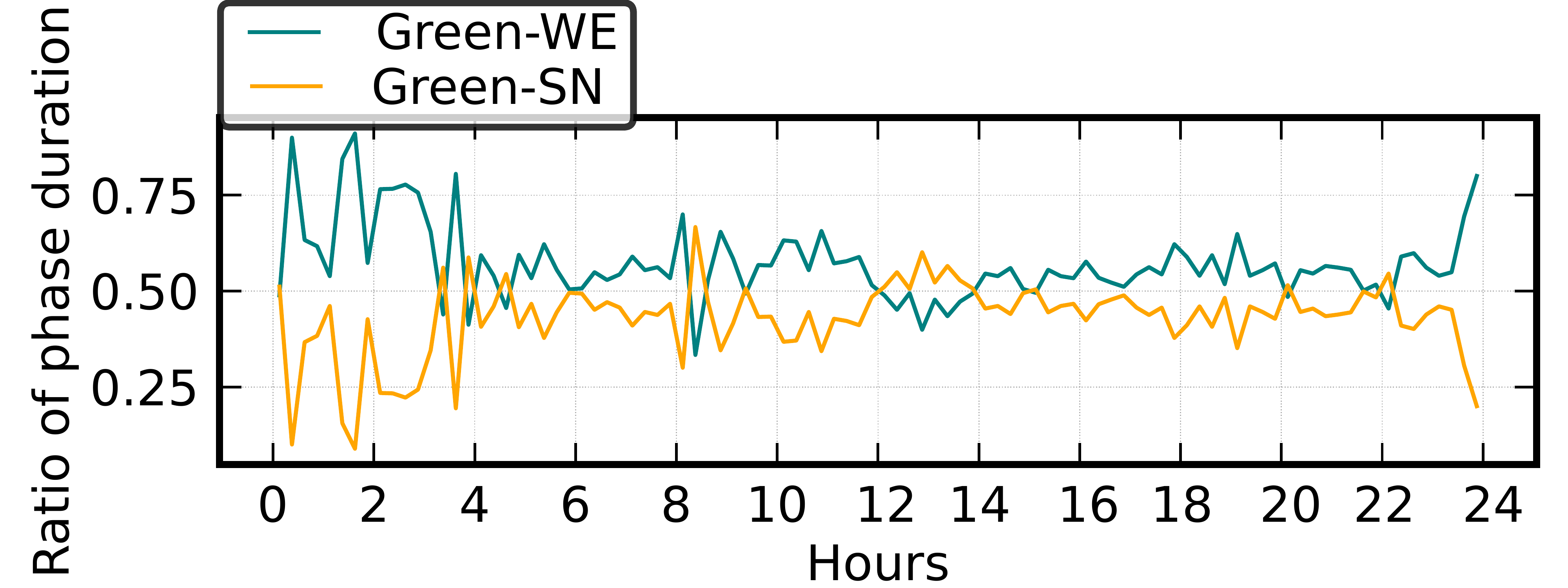} \\
\end{tabular}
\caption{Illustration of Traffic data and model performance on one  intersection from City A. WE-EW denotes total vehicles comming from west and east and Green-WE means green lights on west and east, so do SN-NS and Green-SN.}
\label{fig:3_links_analysis}
\end{figure}

We next conduct a case study to gain additional insight about the performance of \deeplightQueue. Figure~\ref{fig:3_links_analysis} depicts one day's traffic condition as well as the model performance on one intersection from City A. The traffic in this intersection has medium volume and clear morning and evening peaks. In Figure~\ref{fig:3_links_analysis}, arrival rate and travel time of vehicles, and the phase ratio of the proposed model \deeplightQueue are displayed.
For clarity, the vehicles coming from the same direction are merged, so does the phase signal.

From Figure~\ref{fig:3_links_analysis}, we can observe that travel time increases for both \fixedtimeHistory and \deeplightQueue during the morning peak hour. However, \deeplightQueue is more robust to the peak hour traffic and manages to keep the average travel time under 100 seconds. Further, during the evening peak hour, we observe the travel time of \deeplightQueue stays low, suggesting that it has learned how to deal with such peak hour traffic.

\subsection{Variations of State and Reward Design}

\begin{table}[h]
\centering
\caption{Performance of \deeplightQueue with different state and reward combinations on uniform traffic. Current phase $p$ is included in state by default. $\averageWaitingTime$ is average waiting time, $\queueLength$ is queue length, $\numOfVehicles$ is number of vehicles, $\image$ is image, $\delay$ is delay.
Ratio for reward factors are selected through cross validation. Travel time is \textit{31.66} when $\numOfVehicles$ and $\queueLength$ are selected for states and reward, respectively.}
\label{tab:different-uniform}
\begin{tabular}{lc}\toprule
\multicolumn{2}{c}{Reward=$\queueLength$}\\\midrule
State &Travel Time\\\midrule
$\averageWaitingTime$&59.92\\
$\averageWaitingTime, \numOfVehicles$&39.35\\\midrule
$\queueLength$&50.52\\
$\queueLength,\numOfVehicles$&37.93\\\midrule
$\image$&38.16\\
$\image,\numOfVehicles$&38.27\\\bottomrule
\end{tabular}
\quad
\begin{tabular}{lc}\toprule
\multicolumn{2}{c}{State=$\numOfVehicles$}\\\midrule
Reward&Travel Time\\\midrule
$\delay$&105.57\\
$\delay,\queueLength$&40.02\\
\midrule
$\averageWaitingTime$&37.32\\
$\averageWaitingTime,\queueLength$&34.85\\\midrule
$\numOfVehicles$&33.21\\
$\numOfVehicles,\queueLength$&33.46\\\bottomrule
\end{tabular}
\end{table}

In this experiment, we further study the performance of \deeplightQueue under various choices of state and reward (in terms of different combinations of traffic measures). The results are reported in Table~\ref{tab:different-uniform}. We can make the following observations:
\begin{itemize}
\item {\bf State}. Besides the number of vehicles ($\numOfVehicles$), possible choices for state include average waiting time ($\averageWaitingTime$), queue length ($\queueLength$), and image ($\image$). As shown in Table~\ref{tab:different-uniform} (left), including one of these features in the state results in significantly worse performance at the intersections. The travel time decreases when they are supplemented to $\numOfVehicles$, but the performance is alway worse than using $\numOfVehicles$ only (31.66).

\item {\bf Reward}. Similarly, from Table~\ref{tab:different-uniform} (right), we can observe that all the reward factors, i.e., delay $\delay$, average waiting time $\averageWaitingTime$, and number of vehicles $\numOfVehicles$, can neither outperform the results achieved by using $\queueLength$ only when they are utilized alone, or improve the results when they are utilized together with $\queueLength$.
\end{itemize}
The above results are consistent with our previous claim that adopting $\numOfVehicles$ for state and $\queueLength$ for reward is sufficient for the RL agent to find the optimal policy under uniform traffic.

\subsection{Analysis of Three Traits of RL Approach}

\begin{figure}[ht]
\centering
 \includegraphics[width=0.38\textwidth]{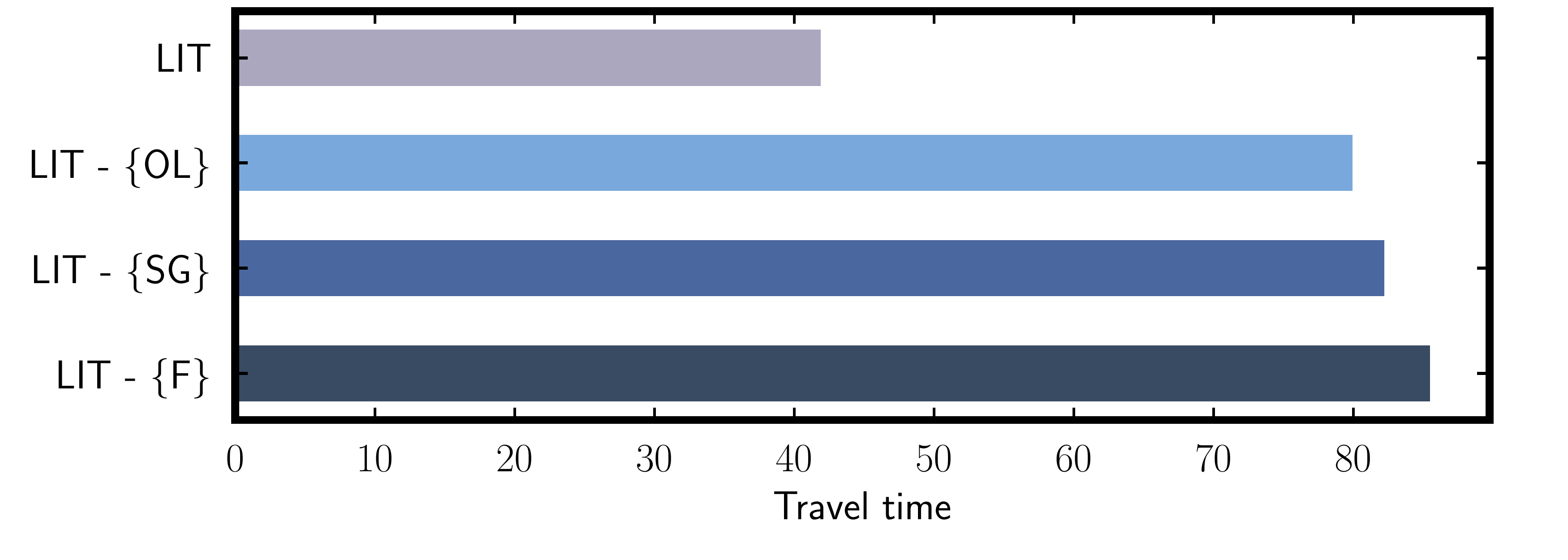}
 \vspace{-3mm}
\caption{Performance of different versions of \deeplightQueue with one component excluded in City A. OL, SG, and F denote online learning, sampling guidance, and forecast, resptively.}
\label{fig:3-trait}
\end{figure}

In Section~\ref{sec:method}, we discuss three important aspects of RL: online learning, sampling guidance, and forecast. In this section, we remove each of the three traits from the \deeplightQueue model to evaluate their contribution to the overall performance on traffic data from City A. The experiment settings are as follows:
\begin{itemize}
\item {\bf Online learning}. We derive a variant of \deeplightQueue that is trained offline and not updated in the testing process.
\item {\bf Sampling guidance}. We design a variant of \deeplightQueue which is trained on random logged samples instead of samples logged by RL algorithm. Specifically, in the training stage, this variant makes random decisions following a probability distribution (e.g., $P(a=0)=0.7$ and $P(a=1)=0.3$). We conduct a grid search on this probability distribution and report the best performance.
\item {\bf Forecast}. We design another variant of \deeplightQueue by setting  $\gamma$ to 0 in Bellman equation~\eqref{eq:bellman}. This will prevent the model from considering future rewards.
\end{itemize}
The results are shown in Figure~\ref{fig:3-trait}. We can observe that removing any component from \deeplightQueue will cause a significant performance deterioration. In other words, all these three traits are essential components of an RL model.

\begin{figure}[h]
\centering
 \includegraphics[width=0.38\textwidth]{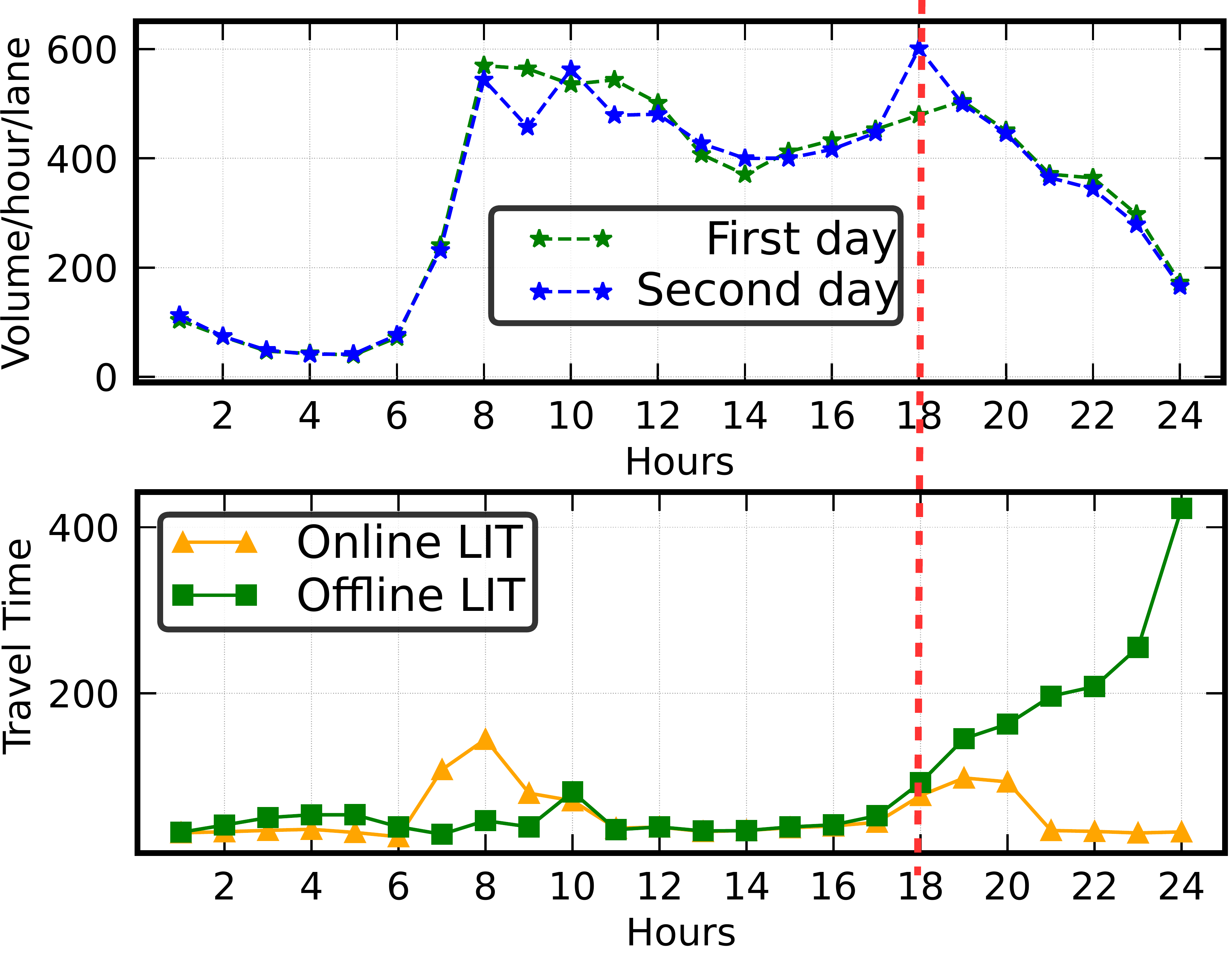}
\caption{Hourly traffic volume and travel time of one intersection with signal controlled by offline/online \deeplightQueue. The online method outperforms the offline one significantly after a sudden traffic increase, which is unseen in the first day.}
\label{fig:oneline_test}
\vspace{-2mm}
\end{figure}

To further illustrate the effect of online learning has on the RL model, we show a case study in Figure~\ref{fig:oneline_test}. Here, the traffic volumes on the two days are very similar. We use the first day for training and the second day for testing.
As shown in Figure~\ref{fig:oneline_test}, both models perform well before 19:00. However, the performance of the offline model starts to degrade around 19:00, when an abrupt change in traffic volume occurs. Offline \deeplightQueue ends up with a traffic jam while online \deeplightQueue always keeps the travel time at a relatively low level.


\section{Conclusion}

In this paper, we systematically examine the RL-based approach to traffic signal control. We propose a new, concise reward and state design, and demonstrate its superior performance compared to state-of-the-art methods on both synthetic and real world datasets. To justify the design of our RL method, we draw connections with classic traffic signal control methods under uniform traffic conditions. In addition, our method is also applicable in different complex multi-intersection scenarios. Further, we analyze three essential factors that make RL outperform the other methods. Extensive experiments are conducted to support our analysis.


\section*{Acknowledgements}
The work was supported in part by NSF awards \#1652525, \#1618448, and \#1639150. The views and conclusions contained in this paper are
those of the authors and should not be interpreted as representing
any funding agencies.

\bibliographystyle{ACM-Reference-Format}
\bibliography{proc}
\newpage
\end{document}